\documentclass[journal]{IEEEtran}

\usepackage{amsmath,amssymb}
\usepackage{graphicx}
\usepackage{booktabs}
\usepackage{multirow}
\usepackage{array}
\usepackage{cite}
\usepackage{url}
\usepackage{xcolor}
\usepackage{balance}
\usepackage{tikz}
\usetikzlibrary{arrows.meta,positioning,fit,backgrounds}

\begin{document}

\title{Revisiting Change VQA in Remote Sensing with Structured and Native Multimodal Qwen Models}

\author{Yakoub Bazi,
Mohamad M.~Al Rahhal,
Mansour Zuair,
and Faroun Mohamed%
\thanks{This work was supported by the Ongoing Research Funding Program, King Saud University, Riyadh, Saudi Arabia, under Grant ORF-2026-995. (Corresponding author: Yakoub Bazi.)}
\thanks{Yakoub Bazi, Mansour Zuair, and Faroun Mohamed are with the Computer Engineering Department, College of Computer and Information Sciences, King Saud University, Riyadh 11543, Saudi Arabia (e-mail: ybazi@ksu.edu.sa; zuair@ksu.edu.sa, faroun@ksu.edu.sa).}
\thanks{Mohamad M. Al Rahhal is with the Applied Computer Science Department, College of Applied Computer Science, King Saud University, Riyadh 11543, Saudi Arabia (e-mail: mmalrahhal@ksu.edu.sa).}
}

\maketitle
\begin{abstract}
Change visual question answering (Change VQA) addresses the problem of answering natural-language questions about semantic changes between bi-temporal remote sensing (RS) images. Although vision–language models (VLMs) have recently been studied for temporal RS image understanding, Change VQA remains underexplored in the context of modern multimodal models. In this letter, we revisit the CDVQA benchmark using recent Qwen models under a unified low-rank adaptation (LoRA) setting. We compare Qwen3-VL, which follows a structured vision–language pipeline with multi-depth visual conditioning and a full-attention decoder, with Qwen3.5, a native multimodal model that combines a single-stage alignment with a hybrid decoder backbone. Experimental results on the official CDVQA test splits show that recent VLMs improve over earlier specialized baselines. They further show that performance does not scale monotonically with model size, and that native multimodal models are more effective than structured vision–language pipelines for this task. These findings indicate that tightly integrated multimodal backbones contribute more to performance than scale or explicit multi-depth visual conditioning for language-driven semantic change reasoning in RS imagery.
\end{abstract}

\begin{IEEEkeywords}
Change visual question answering, semantic change understanding, structured multimodal models, native multimodal models, low-rank adaptation.
\end{IEEEkeywords}

\section{Introduction}

RS change detection is a fundamental problem in Earth observation, with broad relevance to applications such as urban growth monitoring, land-use assessment, environmental surveillance, and disaster analysis. In most practical settings, however, change detection systems produce outputs in the form of binary maps or semantic change labels. Although such outputs are useful for visualization and quantitative analysis, they are less natural when users seek direct answers to questions such as \emph{what changed}, \emph{how it changed}, or \emph{where the change occurred}. This limitation has motivated growing interest in language-centered interaction with RS imagery, where natural-language queries provide a more flexible and accessible interface to visual information. Within this context, Change VQA provides a natural formulation by enabling question answering over bi-temporal imagery and thereby moving beyond static map prediction toward semantic, query-driven change understanding \cite{yuan2022cdvqa}.

This direction builds on earlier progress in RS visual question answering (RS-VQA), which introduced language as a flexible interface for querying Earth observation imagery. Initial efforts such as RSVQA \cite{lobry2020rsvqa} demonstrated the feasibility of RS-VQA, while later benchmarks such as HRVQA \cite{li2024hrvqa} and EarthVQA \cite{wang2024earthvqa} expanded the task toward higher-resolution imagery and more demanding reasoning. At the same time, vision--language modeling in RS has developed rapidly, with recent models such as RS-LLaVA \cite{bazi2024rsllava} and recent surveys \cite{bashmal2023language}, \cite{weng2025vlm_rs_survey} highlighting the growing role of generalist multimodal backbones in RS image understanding. However, much of this progress has remained centered on single-image reasoning, whereas temporal change understanding requires question-guided comparison across multiple observations.

To address the temporal setting, Yuan \emph{et al.} introduced the CDVQA benchmark \cite{yuan2022cdvqa}, where questions are posed over co-registered bi-temporal image pairs rather than single scenes. This makes the task substantially more challenging, since the model must not only interpret each scene but also infer the relevant semantic change between them. CDVQA spans multiple reasoning modes, including change existence, semantic transitions, increase/decrease, largest or smallest change, and discretized change ratios. More recent work has extended this line toward grounding, domain generalization, and more interactive change-language modeling \cite{li2024cdqag,ghazaei2025tcssm,deng2026deltavlm}. These developments show that Change VQA is not simply a temporal extension of conventional RS-VQA, but a more demanding task requiring question-conditioned temporal comparison and fine-grained semantic reasoning over multiple observations.

At the same time, multimodal foundation models have progressed rapidly. Recent generalist VLMs can now handle multiple visual inputs, and in some cases even video, within a unified multimodal framework. This makes them a natural setting for revisiting Change VQA. In particular, the Qwen family offers a useful comparison between two different multimodal designs. Qwen3-VL follows a more structured vision--language pipeline with explicit visual conditioning \cite{bai2025qwen3vl}. Qwen3.5, by contrast, is presented as a native vision--language model trained with early fusion on multimodal tokens, leading to a more tightly integrated multimodal design \cite{qwen35blog2026}. This raises an important question for the RS community: can such modern generalist VLMs address Change VQA effectively without relying on handcrafted change-specific modules?

Motivated by this question, this letter revisits the CDVQA benchmark through a controlled comparison of multimodal formulations for Change VQA. Using recent Qwen-family VLMs, we compare two representative designs: a structured pipeline and a more tightly integrated multimodal formulation across multiple model scales. Rather than proposing another handcrafted model for change analysis, our goal is to understand how well current generalist VLMs can handle this task and how multimodal design influences performance. The study therefore provides an updated empirical view of CDVQA and a clearer picture of what modern generalist VLMs can offer for language-driven change analysis.

\begin{figure*}[t]
    \centering
    \includegraphics[width=\textwidth, trim={0 0.7cm 0 0.2cm}, clip]{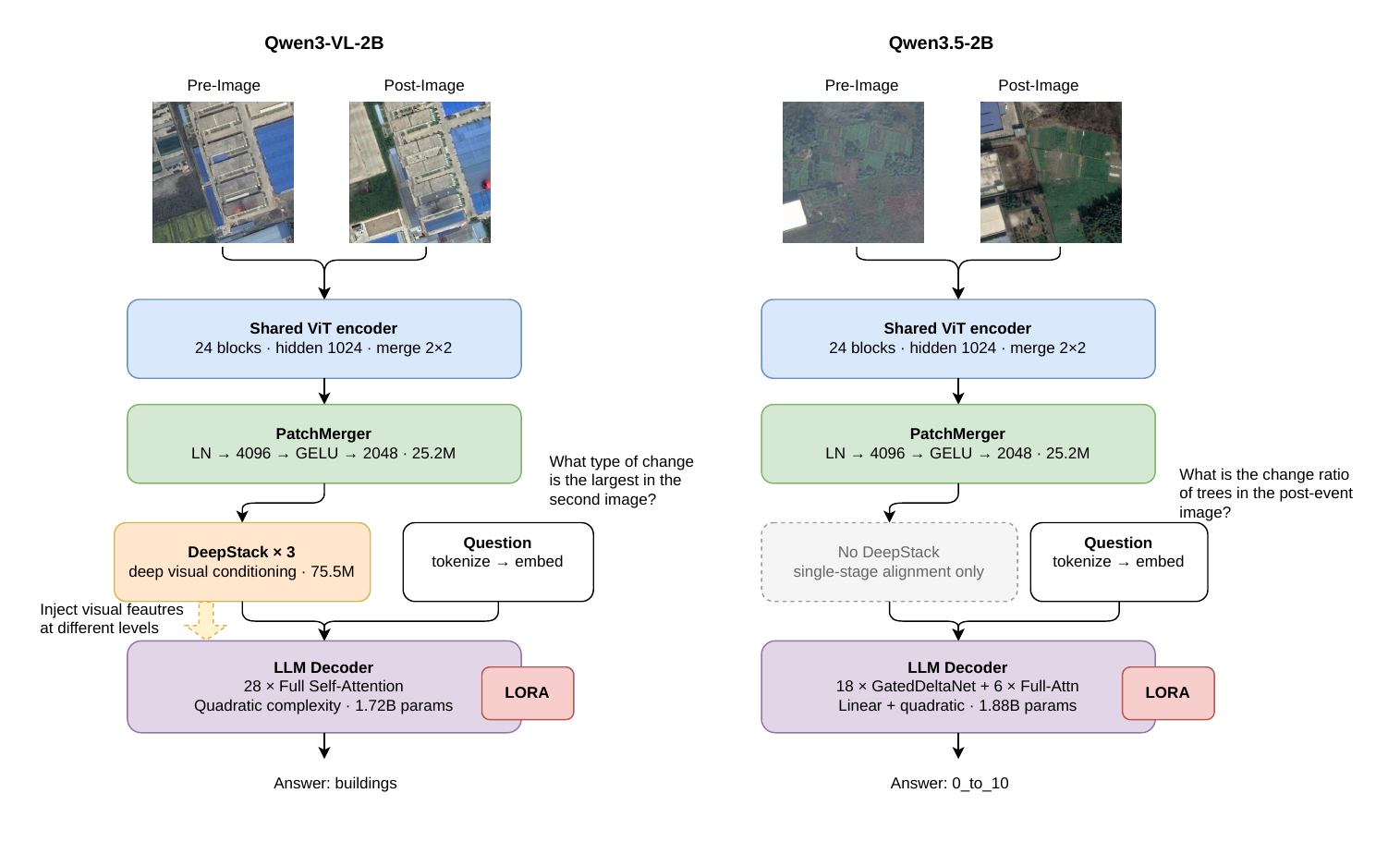}
    \caption{Comparison of the two Qwen-family multimodal formulations used for Change VQA, shown with the 2B variants. Qwen3-VL-2B follows a more structured vision--language pipeline with DeepStack-based multi-depth conditioning and a full self-attention decoder, whereas Qwen3.5-2B adopts a more native multimodal design with a simpler alignment stage and a hybrid decoder based on GatedDeltaNet and full attention.}
    \label{fig:qwen_combined}
\end{figure*}

\section{Methodology}
\subsection{Problem Formulation}
Let $\mathbf{I}^{(1)}\in\mathbb{R}^{H \times W \times 3}$ and
$\mathbf{I}^{(2)}\in\mathbb{R}^{H \times W \times 3}$ denote a co-registered
bi-temporal RS image pair acquired over the same geographic region at two
different times, where $H$ and $W$ denote the image height and width,
respectively. Let $\mathbf{q}=(q_1,q_2,\ldots,q_{L_q})$ denote a natural-language
question of length $L_q$. In Change VQA, the answer is determined by the
question-conditioned semantic relationship between the two temporal
observations. The task therefore requires joint reasoning over
$\mathbf{I}^{(1)}$, $\mathbf{I}^{(2)}$, and $\mathbf{q}$.

In the CDVQA setting~\cite{yuan2022cdvqa}, valid answers belong to a predefined
closed set $\mathcal{Y}=\{y_1,\ldots,y_C\}$, where $C$ is the number of candidate
answers. The prediction problem is therefore written as
\begin{equation}
y^{*}=\arg\max_{y\in\mathcal{Y}}
p_{\theta}\!\left(y \mid \mathbf{I}^{(1)},\mathbf{I}^{(2)},\mathbf{q}\right),
\label{eq:closedset_cvqa}
\end{equation}
where $\theta$ denotes the model parameters.

Unlike specialized Change VQA methods that typically formulate prediction as
classification over a predefined answer space, in this work the task is
formulated within an autoregressive multimodal generation framework. That is,
the answer is modeled as a conditional token sequence even though evaluation is
ultimately performed in a closed-set manner. For an answer sequence
$\mathbf{a}=(a_1,\ldots,a_{L_a})$ of length $L_a$, the underlying conditional
distribution is
\begin{equation}
p_{\theta}\!\left(\mathbf{a}\mid \mathbf{I}^{(1)},\mathbf{I}^{(2)},\mathbf{q}\right)
=
\prod_{t=1}^{L_a}
p_{\theta}\!\left(a_t \mid \mathbf{I}^{(1)},\mathbf{I}^{(2)},\mathbf{q},a_{<t}\right),
\label{eq:ar_factorization}
\end{equation}
which reduces to \eqref{eq:closedset_cvqa} once generation is constrained to
the valid answer set $\mathcal{Y}$ during inference.

\subsection{Multimodal Formulations of the Qwen Families}
Both Qwen families considered in this work receive the same three inputs: a pre-event image, a post-event image, and a question. In both cases, each temporal image is first processed by a vision encoder, and the resulting visual features are mapped into the language-model hidden space before being combined with the question embeddings. The joint multimodal sequence is written as
\begin{equation}
\mathbf{X}=\bigl[\mathbf{V}^{(1)};\mathbf{V}^{(2)};\mathbf{Q}\bigr],
\label{eq:joint_seq}
\end{equation}
where $\mathbf{V}^{(1)}$ and $\mathbf{V}^{(2)}$ denote the merged visual token sequences for the two temporal images and $\mathbf{Q}$ denotes the question embeddings. The comparison in this work is therefore not between different input structures, but between two different multimodal formulations built on the same paired-image and question representation. A conceptual overview is shown in Fig.~\ref{fig:qwen_combined}.

Qwen3-VL follows a more structured vision--language pipeline. After visual token extraction, a PatchMerger projects the visual features into the language-model hidden space, establishing the main visual--language alignment. It then employs additional DeepStack mergers to inject intermediate visual features into the decoder at multiple depths \cite{bai2025qwen3vl}. The resulting interaction is written as
\begin{equation}
\mathbf{H}_{\mathrm{VL}}
=
G_{\mathrm{VL}}\!\left(
\mathbf{X},
\left\{M_{\mathrm{DS}}^{(j)}(\mathbf{Z}_{\mathrm{mid}}^{(j)})\right\}_{j=1}^{3}
\right),
\label{eq:hvl}
\end{equation}
where $\mathbf{Z}_{\mathrm{mid}}^{(j)}$ denotes intermediate visual features, $M_{\mathrm{DS}}^{(j)}$ denotes the $j$th DeepStack merger, and $G_{\mathrm{VL}}$ denotes the multimodal decoder. In this formulation, interaction between the temporal visual tokens and the question is mediated through repeated multi-depth visual conditioning, with decoding performed by a full self-attention backbone.

Qwen3.5 retains a similar visual front-end and an initial PatchMerger that maps visual features into the language space, but it does not use the repeated DeepStack conditioning adopted in Qwen3-VL. Instead, after this single alignment stage, the two temporal views and the question are processed jointly by a more tightly integrated multimodal backbone. In our formulation, this interaction is written as
\begin{equation}
\mathbf{H}_{\mathrm{N}} = G_{\mathrm{N}}(\mathbf{X}),
\label{eq:hn}
\end{equation}
where $G_{\mathrm{N}}$ denotes the hybrid multimodal decoder. Unlike the uniform full-attention decoder used in Qwen3-VL, Qwen3.5 combines GatedDeltaNet and full-attention layers, allowing multimodal information to interact more directly within a unified decoder rather than through repeated explicit conditioning. \cite{qwen35blog2026}.

In essence, the main difference between the two families lies in how multimodal interaction is handled after the initial visual--language alignment. Qwen3-VL follows a more structured pipeline in which visual features are repeatedly reinforced through explicit multi-depth conditioning, whereas Qwen3.5 uses a more integrated multimodal backbone in which the two images and the question interact more directly after the initial alignment step.

\subsection{LoRA Adaptation and Training Objective}
To adapt the Qwen-family models to the Change VQA task efficiently, we employ
Low-Rank Adaptation (LoRA)~\cite{hu2022lora}. Given a pretrained weight matrix
$\mathbf{W}_0 \in \mathbb{R}^{d_{\mathrm{out}}\times d_{\mathrm{in}}}$, LoRA
keeps $\mathbf{W}_0$ frozen and learns a low-rank update
\begin{equation}
\Delta\mathbf{W}=\mathbf{B}\mathbf{A},
\quad
\mathbf{A}\in\mathbb{R}^{r\times d_{\mathrm{in}}},\;
\mathbf{B}\in\mathbb{R}^{d_{\mathrm{out}}\times r},
\end{equation}
so that
\begin{equation}
\mathbf{W} = \mathbf{W}_0 + \frac{\alpha}{r}\mathbf{B}\mathbf{A},
\label{eq:lora_update}
\end{equation}
where $r$ denotes the LoRA rank and $\alpha$ is a scaling factor.

In all experiments, we use the same LoRA hyperparameters ($r=16$, $\alpha=32$)
and apply LoRA to the decoder attention projections (query, key, value, and
output). The vision encoder and visual alignment modules remain frozen.
Note that the resulting number of trainable LoRA parameters is not
identical across model families. This is expected because Qwen3-VL relies on
full self-attention blocks throughout the decoder, while Qwen3.5 uses a hybrid
decoder design with fewer full-attention layers. Under the same target rule
(\emph{Q/K/V/O only}), Qwen3-VL therefore exposes more attention projection
matrices to LoRA than Qwen3.5. For transparency, we report the resulting
trainable parameter counts in Table~\ref{tab:model_scales}.

\begin{table}[t]
\centering
\caption{Model scale and parameter-efficient adaptation details of the evaluated Qwen-family variants.}
\label{tab:model_scales}
\renewcommand{\arraystretch}{1.1}
\begin{tabular}{lccc}
\toprule
\textbf{Model} & \textbf{Params} & \textbf{Trainable (LoRA)} & \textbf{Ratio} \\
\midrule
Qwen3.5-0.8B  & 0.8B & 6.39M  & 0.80\% \\
Qwen3-VL-2B   & 2.1B & 17.43M & 0.83\% \\
Qwen3.5-2B    & 2.2B & 10.91M & 0.50\% \\
Qwen3-VL-4B   & 4B   & 33.03M & 0.83\% \\
Qwen3.5-4B    & 4B   & 21.23M & 0.53\% \\
Qwen3-VL-8B   & 8B   & 66.00M & 0.83\% \\
Qwen3.5-9B    & 9B   & 29.10M & 0.32\% \\
\bottomrule
\end{tabular}
\end{table}

During training, the model is optimized with the standard negative log-likelihood objective
\begin{equation}
\mathcal{L}
=
-\frac{1}{N}
\sum_{i=1}^{N}
\log
p_{\theta}\!\left(
\mathbf{a}_i \mid \mathbf{I}^{(1)}_i,\mathbf{I}^{(2)}_i,\mathbf{q}_i
\right),
\label{eq:training_loss}
\end{equation}
where $\{(\mathbf{I}^{(1)}_i,\mathbf{I}^{(2)}_i,\mathbf{q}_i,\mathbf{a}_i)\}_{i=1}^{N}$ denotes the training set and $\mathbf{a}_i$ is the target answer sequence.

\section{Experiments}

\subsection{Dataset and Experimental Setup}
We evaluate all models on the CDVQA benchmark~\cite{yuan2022cdvqa}, which is designed for change visual question answering on bi-temporal aerial imagery. CDVQA is built from the public subset of the SECOND semantic change detection dataset and contains 2,968 image pairs of size $512 \times 512$ with more than 122,000 automatically generated question--answer pairs. Each sample consists of a pre-event image, a post-event image, and a question, with answers defined over six land-cover classes: non-vegetated ground surface, buildings, playgrounds, water, low vegetation, and trees.

\begin{figure}[t]
    \centering
    \includegraphics[width=\columnwidth]{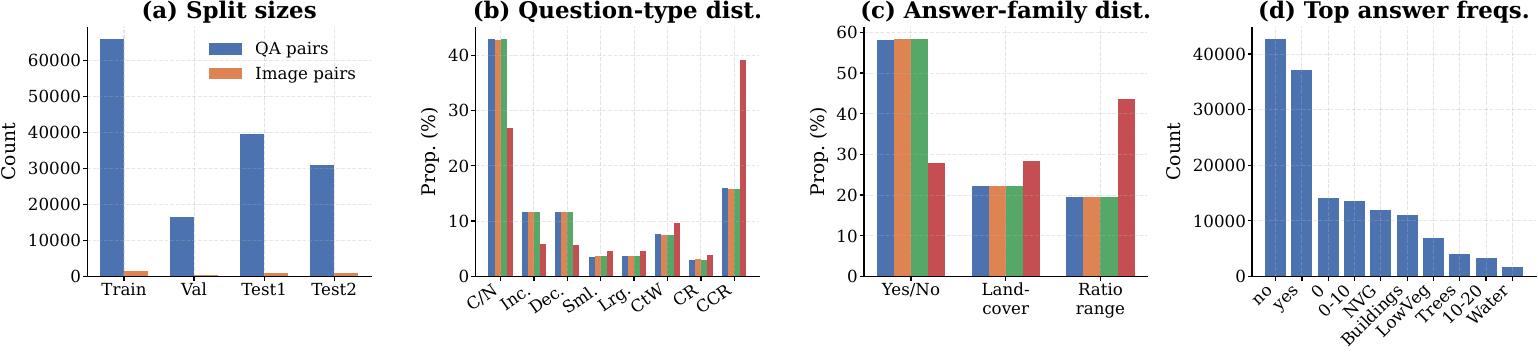}
    \caption{Statistics of the CDVQA benchmark. (a) Number of QA pairs and image pairs in each split. (b) Distribution of question categories across train, validation, test1, and test2. (c) Distribution of answer families. (d) Top answer frequencies over all splits.}
    \label{fig:cdvqa_stats}
\end{figure}

The benchmark is divided into training, validation, and two official test splits. The training split contains 65,967 question--answer pairs from 1,600 image pairs, while the validation split contains 16,441 question--answer pairs from 400 image pairs. The remaining 968 image pairs form two official test sets: test1, with 39,686 question--answer pairs, and test2, with 31,036 question--answer pairs. Following~\cite{yuan2022cdvqa}, we report per-question-type accuracy together with average accuracy (AA) and overall accuracy (OA). The evaluated question types include \emph{change ratio}, \emph{class change ratio}, \emph{change or not}, \emph{change to what}, \emph{increase or not}, \emph{decrease or not}, \emph{smallest change}, and \emph{largest change}. As shown in Fig.~\ref{fig:cdvqa_stats}, the benchmark covers multiple reasoning modes and exhibits a non-uniform distribution across question categories and answers.

All models are fine-tuned on the official training split using LoRA with rank $r=16$ and scaling factor $\alpha=32$. Fine-tuning is conducted using the LLaMA-Factory framework~\cite{zheng2024llamafactory}. We use a batch size of 32 and train for 3 epochs with a learning rate of $5\times10^{-5}$, cosine decay, and a warmup ratio of 0.1. Weight decay is set to 0.01 and the maximum gradient norm to 1.0. Training is performed in bfloat16 with a maximum sequence length of 1024. The best checkpoint is selected on the validation set and then used for testing. Inference time is reported as the average runtime per sample on a single NVIDIA RTX A6000 GPU.

\subsection{Results on the Test Sets}

Table~\ref{tab:main_results} summarizes the CDVQA results of the Qwen variants evaluated in this work. Overall, Qwen3.5 generally performs better than Qwen3-VL at a similar model scale. This difference is especially clear on test2, where the 2B variant improves OA from 65.38 to 70.94. Increasing model size does not always lead to better performance. For example, within Qwen3-VL, the 8B model performs slightly worse than the 4B model on both test sets. A similar trend can be observed for Qwen3.5, where the best overall results are achieved by the 2B variant, while larger models are slower and do not provide consistent gains on this benchmark.

To better understand these results, Table~\ref{tab:main_results} also reports performance for each question type. The binary directional categories, including \emph{change or not}, \emph{increase or not}, and \emph{decrease or not}, remain relatively stable across model variants, particularly on test1. This suggests that the models can reliably capture coarse change existence and direction after adaptation. Larger differences appear in question types that require identifying the dominant semantic pattern. This is most evident on test2 for \emph{largest change} and \emph{change to what}, where Qwen3.5-2B achieves the best results among the compared models.

In contrast, \emph{smallest change} is the most difficult category for all variants. This is likely because it requires fine-grained ranking under subtle differences and very small change regions. \emph{Change ratio} is also challenging, which is expected since estimating discretized change magnitude can easily lead to confusion between neighboring categories. These two question types contribute most of the remaining errors.

Finally, the inference-time results show a clear increase in runtime for larger models. Since the accuracy improvements from scaling are limited, Qwen3.5-2B provides the best trade-off between accuracy and efficiency among the evaluated variants. For this reason, it is used as the representative setting in the following comparison with prior CDVQA methods.

\begin{table*}[t]
\centering
\small
\caption{Per-question-type results of different Qwen-family models on the two official CDVQA test splits (\%). AA and OA denote average accuracy and overall accuracy, respectively.}
\label{tab:main_results}
\resizebox{\textwidth}{!}{
\begin{tabular}{lccccccc|ccccccc}
\toprule
\multirow{2}{*}{Question Type}
& \multicolumn{7}{c|}{\textbf{Test1}}
& \multicolumn{7}{c}{\textbf{Test2}} \\
\cmidrule(lr){2-8} \cmidrule(lr){9-15}
& VL-2B & VL-4B & VL-8B & 3.5-0.8B & 3.5-2B & 3.5-4B & 3.5-9B
& VL-2B & VL-4B & VL-8B & 3.5-0.8B & 3.5-2B & 3.5-4B & 3.5-9B \\
\midrule
change ratio
& 53.88 & 56.78 & 54.45 & 54.01 & 57.55 & 57.15 & 55.61
& 54.06 & 59.48 & 58.47 & 58.20 & 59.57 & 60.41 & 58.95 \\
class change ratio
& 79.12 & 78.79 & 79.31 & 78.47 & 79.86 & 79.08 & 78.83
& 77.04 & 79.21 & 78.70 & 78.65 & 80.68 & 79.45 & 79.50 \\
change or not
& 83.53 & 83.32 & 83.27 & 83.11 & 83.95 & 82.79 & 83.18
& 77.06 & 79.10 & 78.33 & 78.31 & 79.90 & 78.77 & 78.59 \\
change to what
& 61.45 & 61.22 & 59.14 & 61.85 & 61.72 & 61.85 & 61.85
& 59.91 & 61.42 & 59.31 & 61.22 & 64.62 & 61.69 & 62.22 \\
increase or not
& 81.63 & 82.87 & 81.65 & 81.61 & 83.24 & 82.39 & 82.41
& 77.06 & 84.04 & 81.27 & 81.77 & 80.99 & 82.49 & 82.88 \\
decrease or not
& 82.37 & 82.46 & 82.50 & 82.25 & 82.70 & 83.34 & 81.75
& 74.06 & 81.41 & 82.16 & 82.21 & 84.85 & 84.15 & 83.01 \\
smallest change
& 34.09 & 35.26 & 32.33 & 33.57 & 35.23 & 34.78 & 34.64
& 27.89 & 35.64 & 32.16 & 33.92 & 37.22 & 34.78 & 34.30 \\
largest change
& 63.57 & 62.16 & 62.84 & 62.26 & 64.46 & 65.39 & 65.50
& 56.82 & 62.29 & 62.84 & 61.60 & 68.61 & 65.22 & 65.36 \\
\midrule
AA
& 67.45 & 67.86 & 66.94 & 67.14 & 68.59 & 68.35 & 67.97
& 62.99 & 67.82 & 66.65 & 66.99 & 69.56 & 68.37 & 68.10 \\
OA
& 73.85 & 74.08 & 73.47 & 73.53 & 74.74 & 74.22 & 74.05
& 65.38 & 69.33 & 68.29 & 68.50 & 70.94 & 69.71 & 69.44 \\
Inf. time (s/sample)
& 0.12 & 0.17 & 0.35 & 0.13 & 0.16 & 0.28 & 0.37
& 0.16 & 0.18 & 0.35 & 0.13 & 0.16 & 0.28 & 0.37 \\
\bottomrule
\end{tabular}
}
\end{table*}

%
%

\subsection{Comparison with Prior Methods}

To compare our results with earlier CDVQA methods, Table~\ref{tab:sota_compare} reports the performance of the best compact Qwen variant against representative approaches previously evaluated on this benchmark, including the original CDVQA model~\cite{yuan2022cdvqa}, SOBA~\cite{wang2024kernel}, and the recent VisTA framework~\cite{li2024cdqag}. These methods represent the main reference points in the literature, from the original change-aware baseline to more recent solutions.

\begin{table}[t]
\centering
\small
\caption{Comparison with prior methods on the two CDVQA test sets (\%).}
\label{tab:sota_compare}
\resizebox{\columnwidth}{!}{
\begin{tabular}{lcccc}
\toprule
Model & Test1 AA & Test1 OA & Test2 AA & Test2 OA \\
\midrule
CDVQA~\cite{yuan2022cdvqa} & 55.3 & 65.9 & 55.4 & 61.1 \\
SOBA~\cite{wang2024kernel} & 60.3 & 69.2 & 60.3 & 64.8 \\
VisTA~\cite{li2024cdqag} & 65.9 & 73.1 & 65.9 & 68.5 \\
\midrule
Qwen3.5-2B & \textbf{68.59} & \textbf{74.74} & \textbf{69.56} & \textbf{70.94} \\
\bottomrule
\end{tabular}
}
\end{table}

Table~\ref{tab:sota_compare} shows that Qwen3.5-2B outperforms all previous methods on both test sets. In particular, it achieves better AA and OA than the strongest prior model, VisTA, on both test1 and test2. This is an encouraging result because Qwen3.5-2B is a compact general-purpose multimodal model rather than a method specifically designed for change question answering. Overall, this comparison suggests that recent generalist vision-language models can already provide a strong foundation for Change VQA, even without heavily specialized task-specific design.

\begin{figure}[t]
    \centering
    \includegraphics[width=1.05\columnwidth,trim=0 1.2cm 0 0.8cm,clip]{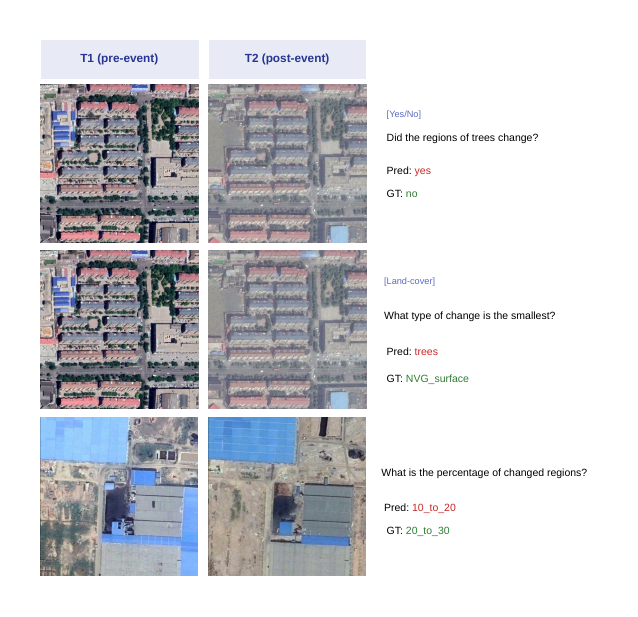}
    \caption{Representative failure cases of Qwen3.5-2B on CDVQA. The examples illustrate three common error modes: false positives under subtle temporal variation, mistakes in fine-grained semantic comparison, and inaccuracies in quantitative change estimation.}
    \label{fig:failure_cases}
\end{figure}

Fig.~\ref{fig:failure_cases} shows representative failure cases of Qwen3.5-2B on CDVQA. Three main error patterns can be observed. First, the model may predict a change when the temporal difference is subtle and the correct answer is negative. Second, it may struggle with fine-grained comparison, especially when distinguishing the smallest semantic change among similar categories. Third, quantitative reasoning remains difficult, as the model may predict a nearby ratio range instead of the correct one. These examples show that, despite its strong overall results, the model still has difficulty with subtle temporal differences, fine-grained semantic reasoning, and precise change estimation.

\section{Conclusion}
This letter revisited Change VQA in RS using modern VLMs on the CDVQA benchmark. The results show that recent general-purpose VLMs can improve performance on this task compared with earlier specialized methods. The results also suggest that native multimodal models are more effective than structured vision--language pipelines for Change VQA, highlighting the importance of tighter interaction between the temporal images and the question. Some challenges still remain, especially in subtle temporal comparison, fine-grained semantic discrimination, and quantitative change estimation. These results suggest that further gains may come from better instruction tuning, stronger reasoning during training, and better question-guided use of the two temporal images. It may also be useful to build on these modern VLMs with lightweight change-aware components, such as change spotting rather than designing entirely new architectures from scratch.
\balance
\bibliographystyle{IEEEtran}
\bibliography{refs}

\end{document}